%% file: root.tex
\documentclass[letterpaper, 10 pt, conference]{ieeeconf}  %

\IEEEoverridecommandlockouts                              %

\usepackage{amssymb}
\usepackage{subfiles}
\usepackage{amssymb}
\usepackage{mathtools}

\usepackage{amsthm}
\usepackage{amsmath}
\let\labelindent\relax
\usepackage{enumitem}
\usepackage{algcompatible}
\usepackage{listings}
\usepackage{algorithm}
\usepackage{booktabs}
\usepackage{subcaption}
\usepackage{graphicx}
\usepackage{xcolor}
\usepackage{wrapfig}
\usepackage{titlesec}
\usepackage{hyperref}
\usepackage{balance}

\makeatletter
\algtext*{ENDIF}
\algtext*{ENDFOR}
\algtext*{ENDWHILE}
\makeatother

\usepackage[backend=biber,style=ieee,natbib=true, url=false,eprint=false,doi=false,citestyle=numeric-comp,maxbibnames=3]{biblatex} 
\definecolor{darkBlue}{RGB}{10,50,220}
\definecolor{customRed}{RGB}{190,110,113}
\definecolor{customGreen}{RGB}{70,170,80}

\hypersetup{
	colorlinks=true,
	linkcolor=customGreen,
	filecolor=magenta,      
	urlcolor=cyan,
	citecolor=customRed,
}
\usepackage[capitalize,noabbrev]{cleveref}
\usepackage{stfloats}  
\title{\LARGE \bf
CLAM: Continuous Latent Action Models\\for Robot Learning from Unlabeled Demonstrations
}

\author{Anthony Liang$^{1,*}$, Pavel Czempin$^{1,*}$, Matthew M. Hong$^{1}$, Yutai Zhou$^{1}$, Jingzhen Wang$^{1}$,\\Erdem Bıyık$^{1,2,\dagger}$, Stephen Tu$^{1,2,\dagger}$ \\
  \texttt{\{anthony.liang,czempin\}@usc.edu}%
\thanks{*Equal contribution}%
\thanks{$\dagger$Equal advising}%
\thanks{$^{1}$Thomas Lord Department of Computer Science, University of Southern California}%
\thanks{$^{2}$Ming Hsieh Department of Electrical and Computer Engineering, University of Southern California}%
}

\input{commands}

\input{macros}

\begin{document}

\maketitle
\thispagestyle{empty}
\pagestyle{empty}

\begin{abstract}
\subfile{sections/00_abstract}

\end{abstract}

\section{Introduction}
\subfile{sections/01_intro}

\section{Related Work}
\subfile{sections/02_rw}

\section{Problem Setting}
\subfile{sections/03_problem}

\section{Continuous Latent Action Models}
\subfile{sections/04_method}

\section{Experimental Setup}
\subfile{sections/05_experiment_setup}

\section{Results}
\subfile{sections/06_results}

\subfile{sections/06.2_ablations}

\subfile{sections/07_conclusion}
\subfile{sections/08_limitations}

\newpage

\section*{APPENDIX}
\subfile{sections/appendix/appendix}

\balance
\printbibliography

\end{document}

%% file: commands.tex
\theoremstyle{definition}

\newcommand{\calL}{\mathcal{L}}
\newcommand{\calD}{\mathcal{D}}

\numberwithin{equation}{section}

\DeclarePairedDelimiterX{\infdivx}[2]{(}{)}{%
  #1\;\delimsize\|\;#2%
}

%% file: macros.tex
\definecolor{blue}{HTML}{4E79A7}
\definecolor{orange}{HTML}{F28E2B}
\definecolor{purple}{HTML}{8E6C8A}
\definecolor{green}{HTML}{59A14F}
\definecolor{red}{HTML}{E15759}
\definecolor{yellow}{HTML}{EDC949}
\definecolor{gray}{HTML}{79706E}
\definecolor{pink}{HTML}{E28A9A}
\definecolor{brown}{HTML}{9C755F}

\definecolor{idm}{HTML}{00B761}
\definecolor{fdm}{HTML}{4878CF}
\definecolor{ad}{HTML}{AF58BA}
\definecolor{policy}{HTML}{EDC949}

\newcommand{\idm}[1]{\textcolor{idm}{#1}}
\newcommand{\fdm}[1]{\textcolor{fdm}{#1}}
\newcommand{\ad}[1]{\textcolor{ad}{#1}}
\newcommand{\policy}[1]{\textcolor{policy}{#1}}

\newcommand{\bc}{\textbf{\texttt{\textcolor{red}{BC-AL}}}}
\newcommand{\bcal}{\textbf{\texttt{\textcolor{red}{BC-Action-Labeled}}}}
\newcommand{\lapo}{\textbf{\texttt{\textcolor{pink}{LAPO}}}}
\newcommand{\lapa}{\textbf{\texttt{\textcolor{brown}{LAPA}}}}
\newcommand{\dynamo}{\textbf{\texttt{\textcolor{purple}{DynaMo}}}}
\newcommand{\vpt}{\textbf{\texttt{\textcolor{blue}{VPT}}}}
\newcommand{\mlpclam}{\textbf{\texttt{\textcolor{orange}{MLP-CLAM}}}}
\newcommand{\mlp}{\textbf{\texttt{\textcolor{orange}{MLP}}}}
\newcommand{\tfclam}{\textbf{\texttt{\textcolor{orange}{Transformer-CLAM}}}}

\newcommand{\tf}{\textbf{\texttt{\textcolor{orange}{Transformer}}}}
\newcommand{\stclam}{\textbf{\texttt{\textcolor{orange}{ST-ViViT-CLAM}}}}
\newcommand{\stc}{\textbf{\texttt{\textcolor{orange}{ST-ViViT}}}}
\newcommand{\stcs}{\textbf{\texttt{\textcolor{orange}{ST-CLAM}}}}
\newcommand{\bce}{\textbf{\texttt{\textcolor{yellow}{BC-Expert}}}}
\newcommand{\bces}{\textbf{\texttt{\textcolor{yellow}{BC-E}}}}

\newcommand{\best}[1]{$\mathbf{\textcolor[HTML]{990000}{#1}}$}

\newcommand{\labeleddata}{\cal{D}_{\text{labeled}}}
\newcommand{\unlabeleddata}{\cal{D}_{\text{unlabeled}}}
\newcommand{\expertdata}{\cal{D}_{\text{unlabeled-expert}}}
\newcommand{\labeledexpertdata}{\mathcal{D}_{\text{relabeled-expert}}}
\newcommand{\task}[1]{\textsc{#1}}

%% file: sections/00_abstract.tex
Learning robot control policies from demonstrations typically requires action-labeled expert data, which is expensive to collect through teleoperation. 
We study a more practical setting in which expert demonstrations are available only as observation sequences without action labels, and only task-agnostic play data contains actions. 
We introduce continuous latent action models (CLAM), a framework that infers continuous latent actions between consecutive observations using self-supervised dynamics prediction.
To ground these latent actions into executable motor commands, CLAM jointly trains an action decoder using a small amount of task-agnostic play data.
We show that continuous latent actions combined with this joint training are essential for high-dimensional continuous control. 
Across DMControl locomotion, MetaWorld manipulation, and real-world WidowX robot tasks, CLAM improves average task success rates by $2{-}3 \times$ over prior latent-action baselines and approaches behavior cloning trained with privileged expert action labels. 
Our results demonstrate that effective robot policies can be learned from unlabeled demonstrations and deployed on real hardware without collecting expert action-labeled data.
Videos and code are available at \href{https://clamrobot.github.io}{clamrobot.github.io}.

%% file: sections/01_intro.tex
Recent efforts in robotics attempted to tackle the collection of large-scale robot  datasets~\citep{xiong2023robotube, open_x_embodiment_rt_x_2023, black2024pi_0}. 
However, data collection ultimately requires manual teleoperation, which is expensive and time-consuming.
Furthermore, fine-tuning models in situ via teleoperation to address the diversity of real-world environments and tasks is challenging for laypeople.

\begin{figure*}[t]
\centering
\includegraphics[width=\textwidth]{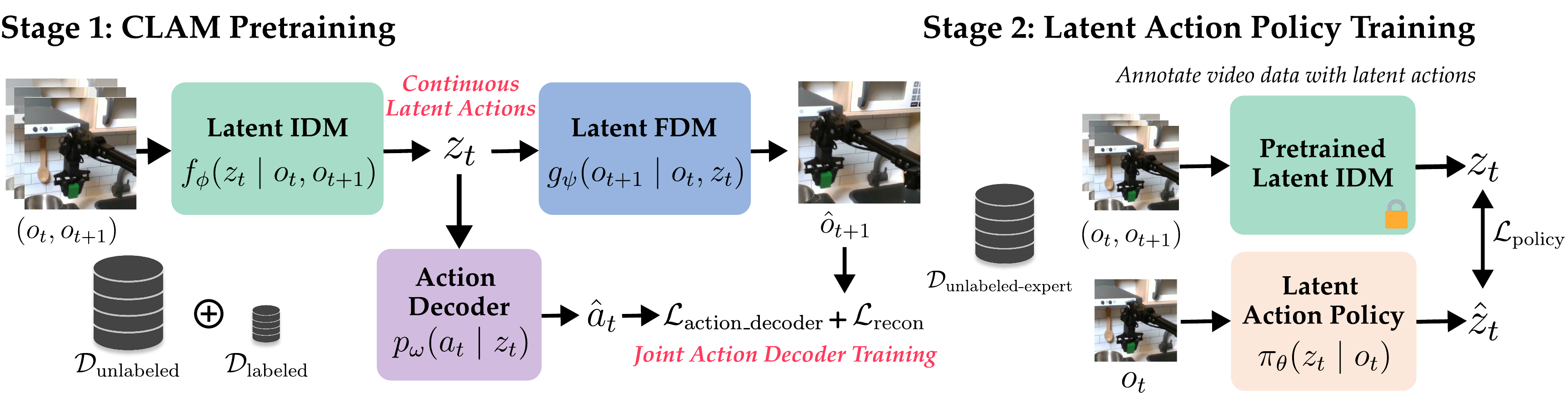}
\caption{\textbf{Overview of CLAM.} CLAM consists of a \textbf{\idm{latent inverse dynamics model}}, $f_\phi$, which infers the latent action between consecutive observations and \textbf{\fdm{latent forward dynamics model}}, $g_\psi$, which predicts the future observation conditioned on the latent action and observation history. 
CLAM learns a latent action space through the self-supervised objective of future observation reconstruction. 
Unlike prior work, CLAM produces \textbf{\emph{continuous latent actions}}. 
To ensure the learned latent space is amenable to decoding to real-world actions, CLAM \textit{\textbf{jointly}} trains the action decoder and the latent action model.}
\label{fig:clam}
\end{figure*}

A promising alternative is training on human video demonstrations.
These are easier to collect than teleoperation data~\cite{hong2025handdatafastrobot} and thus easier to scale to large datasets~\cite{ego4d, Damen2022RESCALING}.
However, utilizing human videos to train robot policies has two main challenges.
First, standard imitation learning (IL) requires demonstrations from the robot's own embodiment and viewpoint.
Recent work addresses this by converting human videos to robot-compatible observations using hand tracking and inpainting ~\cite{lepert2025phantom, lepert2025masquerade}.
However, even with converted video, IL still requires ground-truth robot control signals, which these conversion methods do not provide.
This is precisely the problem we focus on in this paper:
learning from expert demonstrations captured as robot video, but without action labels.

Actionless robot video is abundant in practice: human demonstrations can be converted to robot viewpoints~\cite{lepert2025phantom, lepert2025masquerade}, internet video captures behaviors at scale~\cite{ego4d, Damen2022RESCALING}, and existing robot datasets might contain observation logs without compatible action formats. 
We introduce \emph{continuous latent action models} (CLAM), a framework for learning robot control policies from observation-only robot demonstrations.

CLAM trains a latent inverse dynamics model to infer a continuous latent action between consecutive observations, supervised only by a forward dynamics model that reconstructs the next observation from the current observation and the latent action. 
This self-supervised loop requires no action labels which simplifies training on large-scale data.
To make these latent actions executable on a real robot, CLAM jointly trains an action decoder that maps latent actions to motor commands, using only a small amount of task-agnostic play data. Joint training is critical: it regularizes the latent space so that it remains decodable, which we show is necessary for continuous control. Once trained, the latent inverse dynamics model annotates unlabeled expert demonstrations with pseudo-actions, and a policy is trained via standard IL on the resulting dataset.

However, realizing this pipeline for physical robots requires representations that map seamlessly to continuous robot control signals.
In this work, we identify several key architectural improvements that greatly improve the paradigm of learning from observation-only data in continuous control robotics settings. Our contributions are threefold: 
\begin{enumerate}
    \item We demonstrate that latent action relabeling becomes practical for continuous robot control when using continuous latent spaces and joint grounding.
    \item We demonstrate that task-agnostic play data is sufficient for grounding latent actions, dramatically reducing the cost of data collection for new tasks.
    \item We deploy and evaluate CLAM on a physical WidowX robot across four manipulation tasks, showing it is competitive with privileged expert action labels without ever collecting action-labeled expert demonstrations.
\end{enumerate}

%% file: sections/02_rw.tex
\textbf{Imitation from Observation.} \label{sec:lfo}
Learning robot policies from sequences of observations without access to actions is often referred to as \emph{imitation from observation} (IfO)~\cite{Liu_2018}. 
Unlike standard IL, demonstrations may differ in embodiment, environment, or viewpoint, and do not include action labels. 
Existing IfO approaches either learn policies directly from observation sequences by aligning trajectories~\cite{Liu_2018, yang2019lfo, Liu2020State}, or predict future observations and map them to robot controls using learned or off-the-shelf perception models~\cite{wen2023anypoint, ko2024learning}. 
However, these methods often depend on online rollouts or off-the-shelf vision models, which limit their applicability. 
In contrast, CLAM is trained fully offline, without rollouts or off-the-shelf models.

\textbf{Supervised Learning of Inverse Dynamics Models.} \label{sec:vpt}
With access to some action-labeled data, inverse dynamics models (IDMs) can be trained in a supervised fashion. VPT~\cite{baker2022video} and Seer~\cite{tian2024predictive}, for example, use an IDM to label large unlabeled datasets, while UniPi~\cite{du2023learning} finetunes a video prediction model and applies an IDM to infer robot actions from synthetic data.
A key limitation of such methods is that the IDM must predict actions in the \emph{real} action space.
Consequently, these methods are bottlenecked by the number of labeled demonstrations for task-specific fine-tuning or the budget required to collect them.

\textbf{Latent Action Models.} \label{sec:related-work-lam} 
To bypass the dependency on labeled data, several works learn IDMs in a \emph{latent} action space using only observation data~\cite{Menapace_2021, schmidtlearning, bruce2024genie, cui2024dynamo, ye2024latent, klepach2025object, bu2025learning, chen2025villa0x0}.
These works apply a variety of subtle architectural differences, which we compare in a real-world setup.
Works such as LAPA~\cite{ye2024latent} and DynaMo~\cite{cui2024dynamo} use latent action learning for pretraining to achieve better representation learning performance \cite{ye2024latent, cui2024dynamo, tharwat2025latentactionpretrainingworld}.
However, unlike CLAM, since the latent action model (LAM) is limited to pretraining, each new task still requires collecting labeled expert data to train a policy.
Works such as LAPO~\cite{schmidtlearning} and Genie~\cite{bruce2024genie} combine VPT-style methods with latent action models, using the IDM to annotate unlabeled trajectories directly.
While promising, they focus on video game domains with small discrete action spaces.
These methods use vector quantization in the latent action space; recent works point toward the benefits of using continuous latent action spaces~\cite{nikulin2025latent,yang2025comolearningcontinuouslatent}.
We perform a detailed analysis in the real-world robotics domain, validating key architectural choices that make latent action modeling practical in fully continuous, high-dimensional control settings.

A slew of concurrent works study the applicability of latent action learning to in-the-wild data~\cite{garrido2026learninglatentactionworld}, as a component of large behavior models~\cite{lin2026systematicstudydatamodalities}, and world models~\cite{ye2026worldactionmodelszeroshot, gao2026dreamdojo}, highlighting the importance of this line of research.

LAOM~\citep{nikulin2025latent} found that supervision with a small fraction of labeled actions improves the applicability of a learned latent action space to downstream tasks under distractors.
Unlike LAOM, which used a single linear layer for supervision and discarded this layer at test time, our ground-truth action-supervision comes from the same jointly trained action decoder that is later used at inference time.
Additionally, we focus on simulated and real-world robot manipulation tasks whereas \citet{nikulin2025latent} focus on invariance to distractors in simulated locomotion.

%% file: sections/03_problem.tex
\label{sec:problem}

We consider a practical setting in which a user wants to teach a robot a new task without collecting action-labeled expert demonstrations.
The data regime involves three datasets with complementary roles.
The bulk of the training data comes from a large corpus of robot video collected across diverse environments and tasks, without action labels.
We denote this pool $\unlabeleddata$.
Because no teleoperation or action logging is required, this type of data is cheap to scale.

Separately, we assume access to a small dataset of action-labeled robot transitions, $\labeleddata$. %
In our experiments, this consists largely of undirected, task-agnostic play data, which requires minimal expertise from the operator and no task-specific planning, making it substantially cheaper to collect than expert demonstrations~\cite{lynch2020learning}.
$\labeleddata$ is used solely to ground latent actions into executable motor commands.
Additionally, in our experiments, a small amount of task-specific demonstrations supplement the labeled data, but we assume they cannot be disambiguated from the play data at training time.
Furthermore, we validate the effectiveness of CLAM on purely sub-optimal labeled data in our ablation studies.

The user's only contribution is a small set of task-specific demonstrations, $\expertdata$, collected by teleoperating the robot and performing the target task themselves.  
These demonstrations are expert with respect to the task but contain no action labels and are used to train the final policy.

To isolate the challenge of action inference from the orthogonal challenge of cross-embodiment transfer, we use robot demonstrations with action labels withheld as our unlabeled data.
Additionally, all methods, including CLAM and baselines, have access to the same $\unlabeleddata$, $\labeleddata$, and $\expertdata$, however, some baselines might be bottlenecked by labeled demonstration requirements and thus cannot make use of some of the data.
In this work, we focus on the single-task, single-robot-embodiment setting.

%% file: sections/04_method.tex
\label{sec:method}

We introduce \emph{continuous latent action models} (CLAM), shown in \Cref{fig:clam}, a scalable approach for training continuous control policies from unlabeled observation data.
CLAM consists of two stages.
In {\textbf{\textcolor{orange}{Stage 1}}} (\Cref{sec:idm}), we train a latent action model (LAM) for relabeling observation-only data.
We then use this LAM in {\textbf{\textcolor{orange}{Stage 2}}} (\Cref{sec:la-policy}) to train a latent action policy.

\subsection{Latent Action Model Training}\label{sec:idm}
In {\textbf{\textcolor{orange}{Stage 1}}}, we pretrain a Latent Action Model (LAM) that we later use to annotate trajectories with pseudo-action labels. 
A LAM consists of two components: a \emph{forward dynamics model} (FDM), which predicts the transition dynamics of the environment, and an \emph{inverse dynamics model} (IDM), which inverts this process by inferring the action performed between two subsequent observations.

\begin{algorithm}[tb]
    \caption{CLAM w/ Joint Action Decoder Training}
    \label{alg:clam_joint}
    \begin{algorithmic}[1]
        \STATE \textbf{Input:} $\calD_{\text{unlabeled}}$, $\calD_{\text{labeled}}$, $\calD_{\text{unlabeled\_expert}}$, \textbf{\idm{IDM}} $f_{\phi}$, \textbf{\fdm{FDM}} $g_{\psi}$, \textbf{\ad{Action Decoder}} $p_{\omega}$, \textbf{\policy{Latent A. Policy}} $\pi_{\theta}$
        \Statex $N_C$: number of CLAM update steps
        \Statex $N_{P}$: number of policy updates steps
        \Statex $K$: train action decoder every
        \Statex \textcolor{orange}{\small \textbf{\texttt{\# Stage 1:Train CLAM and Action Decoder}}}
        \FOR{$\text{iter} = 1$ in range($N_C$)}
            \STATE Update $f_\phi$ and $g_\psi$ with $\calL_{\text{recon}}$ on $\calD_{\text{unlabeled}}$
            \IF{iter \% $K$ == 0}
                \STATE Update $p_\omega$ and $f_\phi$ with $\calL_{\text{action\_decoder}}$ on $\calD_{\text{labeled}}$
            \ENDIF
        \ENDFOR
        \Statex \textcolor{orange}{\small \textbf{\texttt{\# Stage 2:Train Latent Action Policy}}}
        \STATE Annotate $\calD_{\text{unlabeled\_expert}}$ with \textbf{\idm{IDM}} $f_{\phi}$
        \FOR{$\text{iter} = 1$ in range($N_{P}$)}
            \STATE Update $\pi_\theta$ with $\calL_{\pi}$ on annotated $\calD_{\text{unlabeled\_expert}}$
        \ENDFOR
    \end{algorithmic}
\end{algorithm}

\begin{algorithm}[H]
    \caption{Inference Time Rollout}
    \label{alg:clam_inference}
    \begin{algorithmic}[1]
        \STATE{\bfseries Input:} \textbf{\ad{Action Decoder}} $p_{\omega}$, \textbf{\policy{Latent Policy}} $\pi_\theta$
        \STATE $o_t = $ env.reset()
        \WHILE{not done}
            \STATE $z_t = \pi_\theta(\cdot \mid o_t)$ \hfill \COMMENT{infer latent action}
            \STATE $a_t = p_\omega(\cdot \mid z_t)$ \hfill \COMMENT{decode latent action}
            \STATE $o_t, \text{done} = $ env.step($a_t$)
        \ENDWHILE
    \end{algorithmic}
\end{algorithm}

Since we train these models without any action labels, we train a \textbf{\idm{\emph{latent} IDM}}, $f_{\phi}(z_t \mid o_t, o_{t+1})$, which predicts an unobserved \emph{latent} action $z_t$ between two consecutive observations.
This $z_t$ encodes transition-specific state-dependent information needed to reconstruct the next observation.
Thus it is a low-level action code and not a high-level skill label.
To provide a training signal for the latent action, we jointly train a \textbf{\fdm{\emph{latent} FDM}}, $g_{\psi}(o_{t+1} \mid o_t, z_t)$, to infer the next observation conditioned on the current observation and latent action.
Since observations are \emph{partial} and do not capture the full environment state, in practice, we provide the LAM with additional $H$ steps of context making it easier to infer the underlying state and predict a more accurate latent action, i.e., $f_{\phi}(z_t \mid o_{t-H},\dots,o_t,o_{t+1})$ and $g_{\psi}(o_{t+1} \mid o_{t-H},\dots,o_t, z_t)$.

As shown in \Cref{fig:clam}, the training signal comes from future observation reconstruction, i.e. $\calL_{\text{recon}} = \text{MSE}(\hat{o}_{t+1}, o_{t+1})$ where $\hat{o}_{t+1}$ is the FDM prediction. 
Our encoder/decoder architecture induces an information bottleneck, ensuring a meaningful, compact action representation rather than shortcut solutions.
While prior works \cite{schmidtlearning, bruce2024genie} discretize latent actions from the IDM using Vector Quantization (VQ)~\cite{van2017neural}, our experiments show that this fails in robotics tasks where actions are inherently continuous. 
We solve this shortcoming by replacing the VQ-based discretized action space with a learned continuous actions.

\textbf{Latent Action Decoder.} 
At test time, the learned latent actions cannot be directly executed in the environment.
Consequently, we learn a \textbf{\ad{latent action decoder}}, $p_\omega(a_t \mid z_t)$ using $\labeleddata$ to ground the learned latent actions to executable environment actions.
To ensure $z_t$ remains a low-level latent action, the action decoder is intentionally not conditioned on observations.
Some prior work \cite{schmidtlearning} trains the action decoder independently from the latent action model. 
This is reasonable for environments with discrete action spaces, where it is possible to learn the mapping from a discrete set of codes to the environment actions.
Recent work \cite{nikulin2025latent} points toward the benefit of providing additional supervision by jointly training with labeled data.
We find this holds in real-world experiments, so we adopt a joint training approach for our LAM training.
We find that this allows for effective decoding to real-world actions.

Importantly, we do not make any assumptions about how $\labeleddata$ is collected. Furthermore, we demonstrate that $\labeleddata$ \textit{can come from any behavioral policy, even a random policy or task-agnostic play data}, allowing CLAM to work even without access to expert teleoperated data.
During LAM pretraining, we alternate between gradient updates on batches of unlabeled data for training the LAM and batches of labeled data for training the action decoder.
The final training objective for CLAM is $\calL_{\text{CLAM}} = \cal{L}_{\text{recon}} + \beta \calL_{\text{action-decoder}}$ where ${\calL_{\text{action-decoder}} = \text{MSE}(\hat{a}_t, a_t)}$ and $\beta$ is a hyperparameter that balances the reconstruction and action decoder losses.

\subsection{Latent Action Policy Training}\label{sec:la-policy}
During {\textbf{\textcolor{orange}{Stage 2}}}, we use the latent IDM from our pretrained CLAM to annotate $\expertdata$ with latent actions. 
The latent FDM only provides the learning signal for training the latent IDM and is discarded at this point. 
We apply the latent IDM to infer the latent action $z_t$ between each consecutive observation $(o_t, o_{t+1})$, i.e., $\labeledexpertdata$ $= \{(o_1^i, z_1^i, o_2^i, z_2^i, \dots, z_{T-1}^i, o_T^i) \,\, \forall \tau^i \in \expertdata\}$. 

Subsequently, we train a \textbf{\policy{latent action policy}}, $\pi_{\theta}(z_t \mid o_t)$, using IL by optimizing $\calL_{\pi} = \text{MSE}(\hat{z}_t,z_t)$ on batches of annotated data from $\labeledexpertdata$. 
During inference time, detailed in \Cref{alg:clam_inference}, our learned policy predicts the latent actions given an observation, which the action decoder will decode into an environment action. 

\textbf{Leveraging pretrained IDM image features for policy training.}
A side-effect of learning a LAM on image-based observations is that the IDM's image encoders can be used as pretrained image features for the latent action policy ({\textbf{\textcolor{orange}{Stage 2}}}). 
Viewed through this lens, training the LAM can be seen as a form of self-supervised representation learning. 
In DynaMo~\cite{cui2024dynamo}, it is shown that pretraining vision encoders using an IDM/FDM self-supervised loss improves the performance of downstream IL from expert-labeled demonstrations. 
We will show in 
\Cref{sec:experiments} that
similar positive transfer also occurs for CLAM when learning latent action policies.

%% file: sections/05_experiment_setup.tex
\label{sec:experiments}

\begin{table*}
\centering
\begin{tabular}{l|cc|cccc|c}
\toprule
 & \task{HalfCheetah} & \task{Hopper} & \task{Assembly} & \task{Bin Picking} & \task{Peg Insert} & \task{Shelf Place} & Average \\
\midrule
\bc & 0.22 {\scriptsize $\pm 0.05$} & 0.35 {\scriptsize $\pm 0.04$} & 0.34 {\scriptsize $\pm 0.05$} & 0.27 {\scriptsize $\pm 0.12$} & 0.29 {\scriptsize $\pm 0.07$} & 0.00 {\scriptsize $\pm 0.00$} & 0.24 \\
\lapo & 0.12 {\scriptsize $\pm 0.05$} & 0.24 {\scriptsize $\pm 0.03$} & 0.15 {\scriptsize $\pm 0.04$} & 0.02 {\scriptsize $\pm 0.03$} & 0.17 {\scriptsize $\pm 0.04$} & 0.06 {\scriptsize $\pm 0.08$} & 0.13 \\
\lapa & 0.22 {\scriptsize $\pm 0.05$} & 0.30 {\scriptsize $\pm 0.06$} & 0.24 {\scriptsize $\pm 0.07$} & 0.15 {\scriptsize $\pm 0.01$} & 0.25 {\scriptsize $\pm 0.02$} & 0.12 {\scriptsize $\pm 0.02$} & 0.21 \\
\dynamo & 0.18 {\scriptsize $\pm 0.03$} & 0.22 {\scriptsize $\pm 0.02$} & 0.10 {\scriptsize $\pm 0.03$} & 0.06 {\scriptsize $\pm 0.03$} & 0.12 {\scriptsize $\pm 0.04$} & 0.08 {\scriptsize $\pm 0.02$} & 0.13 \\
\vpt & 0.32 {\scriptsize $\pm 0.04$} & 0.41 {\scriptsize $\pm 0.03$} & 0.40 {\scriptsize $\pm 0.08$} & 0.05 {\scriptsize $\pm 0.02$} & 0.49 {\scriptsize $\pm 0.06$} & 0.02 {\scriptsize $\pm 0.00$} & 0.28 \\
\mlpclam$^{*}$ & 0.64 {\scriptsize $\pm 0.05$} & 0.64 {\scriptsize $\pm 0.03$} & 0.53 {\scriptsize $\pm 0.04$} & 0.68 {\scriptsize $\pm 0.05$} & 0.58 {\scriptsize $\pm 0.04$} & 0.72 {\scriptsize $\pm 0.04$} & 0.63 \\
\tfclam$^{*}$ & \best{0.72} {\scriptsize \best{\pm 0.04}} & \best{0.81} {\scriptsize \best{\pm 0.05}} & \best{0.91} {\scriptsize \best{\pm 0.03}} & \best{0.82} {\scriptsize \best{\pm 0.03}} & \best{0.79} {\scriptsize \best{\pm 0.07}} & \best{0.93} {\scriptsize \best{\pm 0.02}} & \best{0.83} \\
\bce & 0.68 {\scriptsize $\pm 0.02$} & 0.76 {\scriptsize $\pm 0.04$} & 1.00 {\scriptsize $\pm 0.00$} & 0.94 {\scriptsize $\pm 0.05$} & 0.91 {\scriptsize $\pm 0.03$} & 0.93 {\scriptsize $\pm 0.00$} & 0.87 \\
\bottomrule
\end{tabular}
\caption{\textbf{MetaWorld State-Based  Results}.
We report normalized returns for DMControl tasks and average task success rates for MetaWorld tasks. 
\best{Maroon} highlights the best method in each environment, excluding \bce, which is trained with expert labeled data. 
Our methods, denoted by an asterisk ($*$), outperform all baselines across tasks.}
\label{tab:state_results}
\end{table*}

\begin{figure*}[t]
\centering
\includegraphics[width=0.85\textwidth]{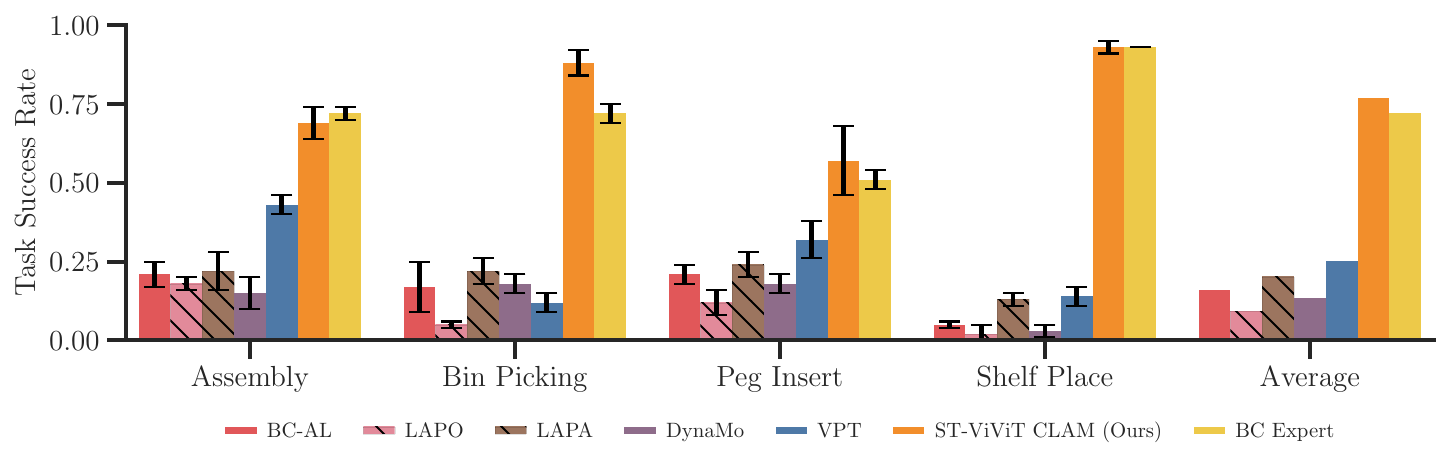}
\caption{\textbf{MetaWorld Image-Based Experiments.} 
We average task success rates over 50 evaluation rollouts across 3 random seeds using image-based inputs. 
We train all methods with the same amount of action-labeled data (100 trajectories) for fair comparison. 
Because $\labeleddata$ is not solely expert data, baselines struggle to learn a performant downstream policy, whereas \mlpclam\ and \tfclam\ achieve up to a $3\times$ improvement over the best baseline. 
Baselines using a \emph{discrete} latent action space are indicated with hashed markers. 
We also report results for \bce, trained on the same number of \emph{labeled expert} trajectories, to illustrate the performance of a privileged reference trained with ground-truth actions.
}
\label{fig:image_results_mw}
\end{figure*}

\textbf{Environments and Datasets.} We compare CLAM to several state-of-the-art baselines using both state- and image-based observations across tasks in DMControl \cite{todorov2012mujoco}, MetaWorld \cite{yu2020meta}, and on a real WidowX robot arm outlined below.
Full details of the data split is provided in \autoref{tab:data_splits}.
\begin{itemize}
    \item \textbf{DMControl.} We evaluate on two DMControl locomotion tasks: \task{Hopper} and \task{HalfCheetah}. 
    We use trajectories from \texttt{medium-expert} split of the D4RL~\cite{fu2020d4rl} benchmark for pretraining CLAM and subsample only suboptimal trajectories for $\labeleddata$.
    \item \textbf{MetaWorld.} We evaluate on four difficult MetaWorld tasks: \task{Assembly}, \task{Bin Picking}, \task{Peg Insert}, and \task{Shelf Place}. We train single-task RL agents and collect the replay buffer data for pretraining. 
    We hold out a separate \texttt{random-medium} dataset similar to DMControl for $\labeleddata$. 
    \item \textbf{WidowX Robot Arm.} We evaluate on four real world manipulation tasks (see bottom of \Cref{fig:real_robot}): \task{Reach Block}, \task{Push Button}, \task{Close Microwave}, and \task{Put Object in Pot and Slide Pot} which requires more than 150 timesteps at a 5 Hz control frequency. 
    Our WidowX robot arm setup is shown in \Cref{fig:real_robot} (Top).
    We collect a task-agnostic, play dataset of $\sim\!50$k transitions for $\unlabeleddata$ and use about $5$k transitions for $\labeleddata$. We separately collect $\sim\!30$ expert demonstrations per task without action labels for $\expertdata$. 
\end{itemize} 
All methods that predict environment actions use each benchmark's native continuous action space.
Ground-truth action spaces differ across benchmarks but not across tasks within a benchmark; see \autoref{tab:action_spaces} for full details.

\textbf{Baselines.} 
To ensure fair comparison, we reuse the same architectural components between CLAM and all baselines.  
These components are an IDM, an FDM, an action decoder/action head, and a transformer BC policy.
Furthermore, all methods have access to same data when applicable.
Some algorithms are unable to leverage certain subsets of the data, e.g. BC cannot use $\mathcal{D}_{\text{unlabeled}}$ since it requires action supervision. 
We compare CLAM to the following baselines: 
\begin{itemize}
    \item \textbf{\bcal\ (\bc)}: Behavior cloning on the small $\labeleddata$, which is not fully expert data. 
Since BC needs action labels, it does not use $\unlabeleddata$.

    \item \textbf{\vpt\ \cite{baker2022video}}: The IDM is trained \textbf{\emph{only}} on $\labeleddata$ via supervised learning.
The IDM is used to label $\expertdata$ with environment actions and a BC policy is then trained on the annotated data. 

    \item \textbf{\lapo\ \cite{schmidtlearning}}: Latent action model with discrete, vector-quantized latent actions.

    \item \textbf{\lapa\ \cite{ye2024latent}}: After LAM pretraining, the final layer of the IDM is replaced with an action head and fine-tuned end-to-end on $\labeleddata$, which is \emph{non-expert}. 

    \item \textbf{\dynamo\ \cite{cui2024dynamo}}: Self-supervised learning on $\unlabeleddata$ to train a vision encoder using an IDM and FDM to predict latent embeddings of future frames. 
    A BC policy is trained on the image embeddings produced by the pretrained vision encoder using $\labeleddata$.
    We include this baseline to compare what a pure image-feature pretraining approach can achieve in our problem setting, which does not use the LAM for relabeling.

    \item \textbf{\mlp-, \tf-, \stc-CLAM (Ours)}: Continuous latent action model with different parameterizations of the latent IDM and FDM.

    \item \textbf{\bce\ (\bces)}: 
    Privileged BC on $\expertdata$ with \emph{ground-truth} action labels available, which are not available to other methods.
\end{itemize}

\textbf{CLAM Model Architectures:} 
For image observations, we model after the Space-Time (ST) Transformer \cite{bertasius2021space}.
We first patchify a $64 \times 64 \times 3$ image with a patch size of 16 for a total of 16 patches. Each patch
is embedded through a linear layer into the hidden dimension. The encoder consists of $N_{E}$ layers of
Space-Time (ST) Attention blocks. 
Each ST block consists of spatial attention followed by temporal
attention and a feedforward layer with skip connection, LayerNorm, and dropout applied between each attention.
The decoder ST block also applies a cross-attention with the latent actions generated by the encoder. 
We add an additional token in the sequence of patch embeddings as a \texttt{CLS} token for the whole image. 
From the \texttt{CLS} token for each timestep, we apply a linear layer to predict the latent actions.

\textbf{Latent Policy Architecture:} 
For our policy, we use a transformer decoder similar to the one used in ACT \cite{zhao2023act}. 
For external and over-the-shoulder RGB images, we use a pretrained ResNet to extract $7 \times 7$ feature maps and flatten them along the spatial dimension to create a sequence of $ d_v$-dimensional tokens, where $d_v$ is the output dimension of the ResNet. 
We use ResNet18 with $d_v = 512$. 
We feed as input to a causal transformer decoder a sequence of learnable action tokens with dimension $d$.
The flattened image feature map is the key and value, and we apply cross-attention between the image features and learnable tokens.
We concatenate all modality tokens and add additional modality-specific embeddings and sinusoidal positional embeddings.

The input sequence to the transformer is a fixed position embedding, with dimensions $k \times 512$ where $k$ is the chunk size and the keys and values are the combined image tokens from the stem. 
At each timestep, we predict a chunk of 5 actions corresponding to 1 second of execution.

%% file: sections/06_results.tex
\label{sec:results}
We aim to study the efficacy of CLAM as a general approach to learn from actionless data, evaluate its ability to train robot policies without access to labeled expert data, and analyze its design choices and limitations. 
We organize our experiments to answer the following:
\begin{enumerate}[label=\textbf{(Q\arabic*)}]
    \item \label{q1} How effective is CLAM at learning policies \emph{without} action-labeled expert demonstrations?
    \item \label{q2} How important is each component of CLAM: \emph{continuous} latent actions and \emph{jointly} training action decoder?
    \item \label{q3} Can CLAM scale to efficiently learn capable robot policies for real-world scenarios?
\end{enumerate}

\begin{figure}[t]
    \centering
    \includegraphics[width=0.7\linewidth]{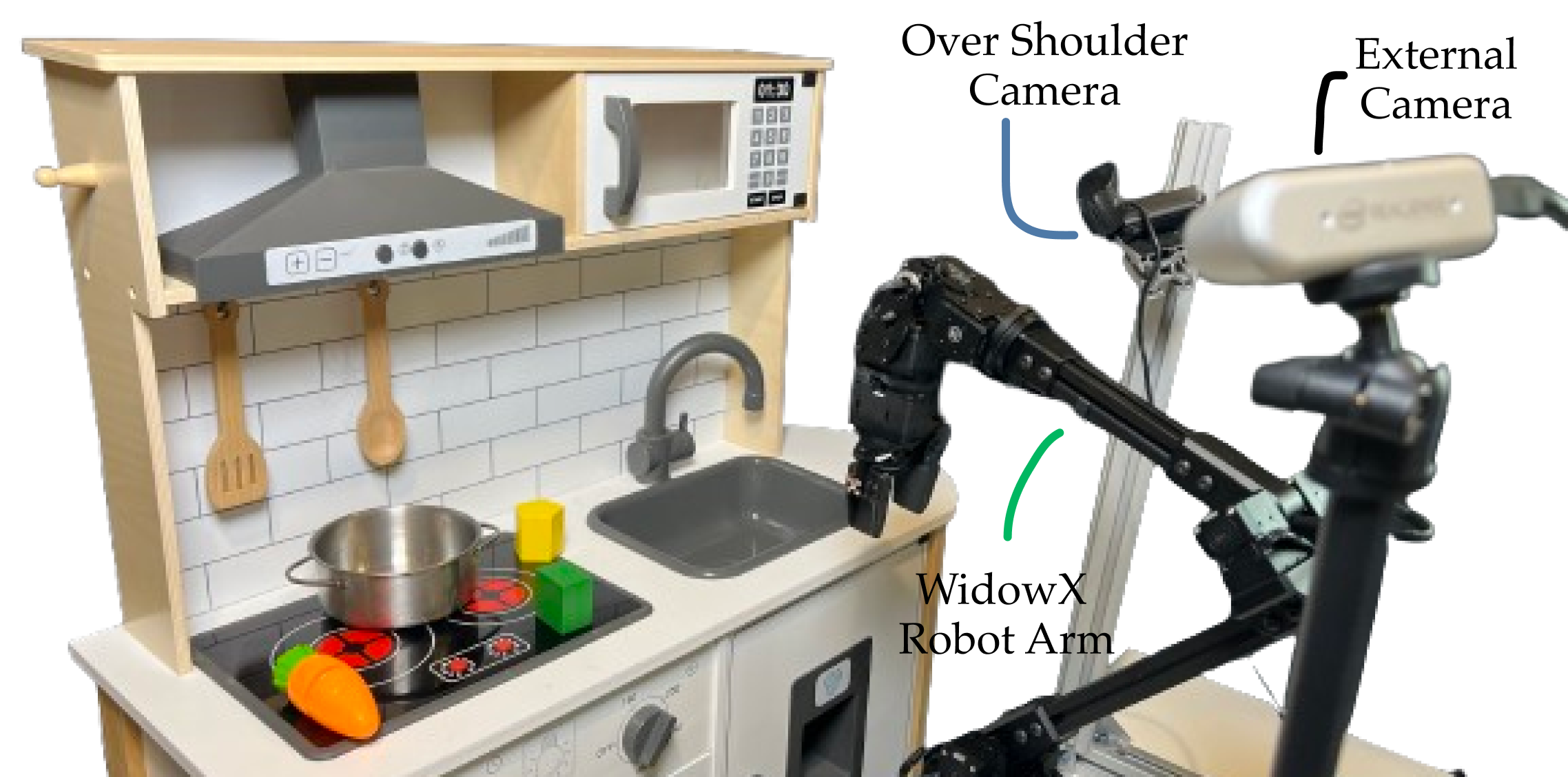}

    \vspace{0.7em} %

    \includegraphics[width=0.6\linewidth]{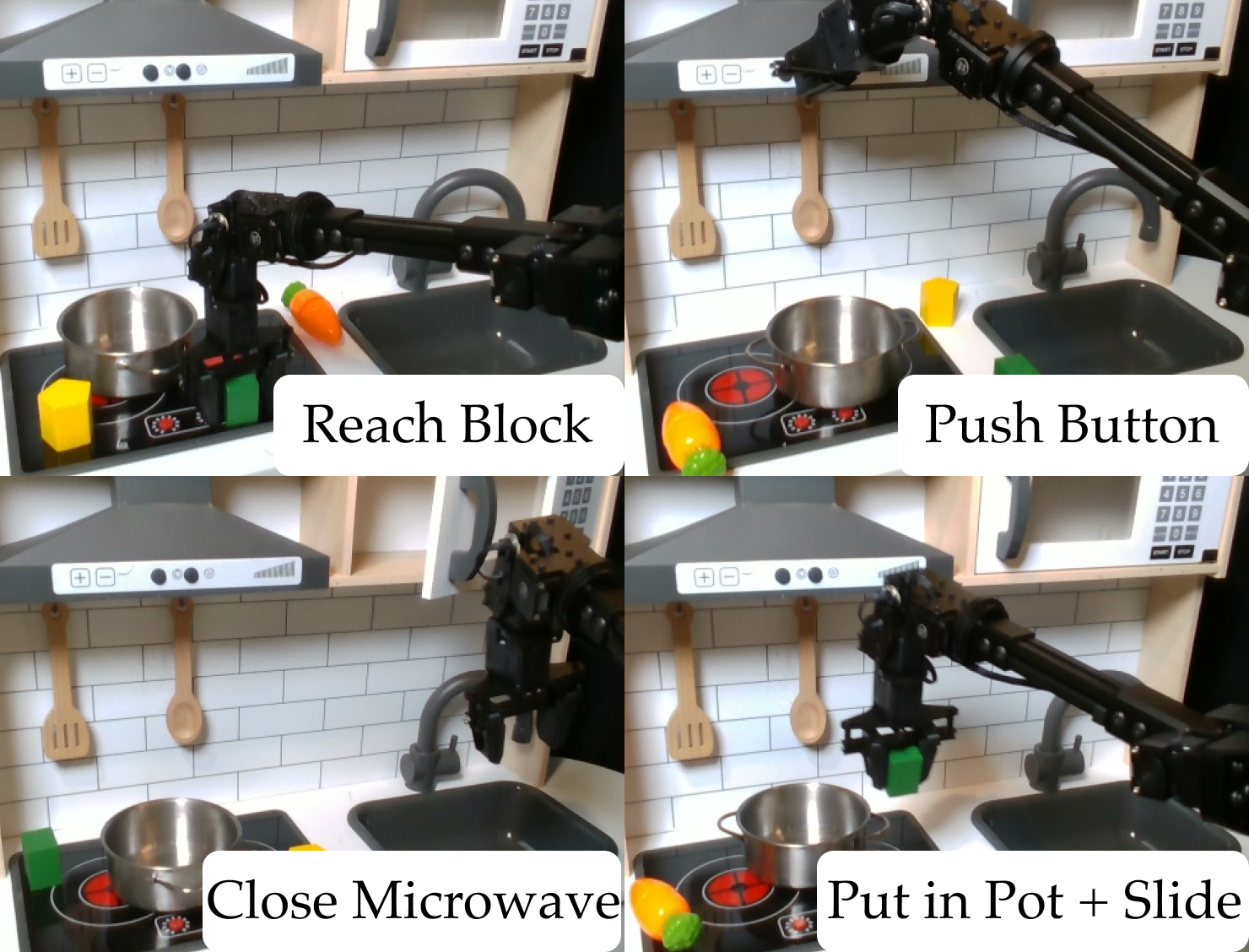}

    \caption{\textbf{WidowX Robot Arm Setup and Evaluation Tasks.} 
    (Top) WidowX robot arm in a toy kitchen setup \cite{walke2023bridgedata}. We use an Intel Realsense D435 RGBD camera as a fixed external camera and a Logitech C920 webcam as an over-the-shoulder camera view. 
    (Bottom) Four tabletop manipulation tasks used to evaluate the scalability of CLAM.}
    \label{fig:real_robot}
\end{figure}

\textbf{\textsc{Finding 1:} 
CLAM outperforms all baselines and nearly matches the performance of BC with expert data in state- and image-based experiments.} 
\Cref{tab:state_results} summarizes our results for state-based inputs on DMControl and \Cref{fig:image_results_mw} for image-based results in MetaWorld. 
CLAM improves upon the best baseline \vpt{} by more than $2 \times$ average normalized return on the DMControl (\textbf{\emph{locomotion}}) tasks and around $2{-}3 \times$ success rate in MetaWorld (\textbf{\emph{manipulation}}).

\bc\ using \emph{action-labeled} data unsurprisingly does not perform well due to imitating suboptimal demonstrations.
In several tasks, \tfclam{} achieves performance close to or even better than that of \bce\, which uses the same amount of privileged \emph{expert} action-labeled data.
Although \bce\ has access to privileged expert action labels, it is trained only on the limited expert dataset.
CLAM, by contrast, uses the larger unlabeled corpus during LAM pretraining before fitting the downstream latent policy. 
Thus, when CLAM approaches or exceeds \bce, we interpret the result as evidence that self-supervised pretraining can compensate for missing expert action labels by learning task-relevant representations and a smoother latent action space. 
In image-based experiments, this benefit is consistent with transfer from the pretrained IDM encoder (cf.~Section~\ref{sec:la-policy}). 
In state-based experiments, the latent-action bottleneck may act as a regularizer
by filtering high-variance action targets. 
We leave a full causal attribution to targeted ablations, but the comparison shows that privileged expert labels are not always sufficient to overcome the data and representation advantages provided by CLAM pretraining.

\begin{table}
\centering
\begin{tabular}{@{}l|c@{\hskip 6pt}c@{\hskip 6pt}c@{\hskip 6pt}c@{}}
\toprule
\textbf{} & \task{Assembly} & \task{Bin Pick} & \task{Peg Insert} & \task{Shelf Place} \\
\midrule
Disc., $\neg$JT & $0.15 \pm \text{\scriptsize 0.03}$ & $0.12 \pm \text{\scriptsize 0.02}$ & $0.18 \pm \text{\scriptsize 0.04}$ & $0.19 \pm \text{\scriptsize 0.03}$ \\
Disc., \; \hspace{1pt}JT & $0.14 \pm \text{\scriptsize 0.04}$ & $0.14 \pm \text{\scriptsize 0.03}$ & $0.17 \pm \text{\scriptsize 0.03}$ & $0.16 \pm \text{\scriptsize 0.04}$ \\
Cont., $\neg$JT & $0.28 \pm \text{\scriptsize 0.04}$ & $0.18 \pm \text{\scriptsize 0.03}$ & $0.21 \pm \text{\scriptsize 0.08}$ & $0.26 \pm \text{\scriptsize 0.06}$ \\
Cont., \; \hspace{1pt}JT & \best{0.69 \pm \text{\scriptsize 0.05}} & \best{0.82 \pm \text{\scriptsize 0.04}} & \best{0.57 \pm \text{\scriptsize 0.11}} & \best{0.88 \pm \text{\scriptsize 0.02}} \\
\bottomrule
\end{tabular}
\caption{\textbf{CLAM Ablation Study.} Both continuous latent actions and joint training are necessary to improve task success. 
Using continuous actions (row~2 $\rightarrow$ row~3) yields about a $1.5\times$ improvement in success rates. 
Joint training (row~3 $\rightarrow$ row~4) further boosts performance by roughly $3\times$. 
JT refers to \emph{joint action decoder training}.}
\label{tab:ablate_vq_joint}
\end{table}

All variants of CLAM outperform the best baseline \vpt~\cite{baker2022video}, highlighting the fact that latent action models scale with $|\unlabeleddata|$ while supervised IDMs only scale with $|\labeleddata|$. 
Since our problem setup assumes $|\labeleddata| \ll |\unlabeleddata|$, it is likely that \vpt\ learns a suboptimal IDM, underscoring the benefit of \emph{latent} action models which can leverage vast, unstructured observation data to learn latent actions in an unsupervised manner. 
CLAM outperforms state-of-the-art methods in our problem setting where only \emph{play} data is available as action-labeled data, and expert data is actionless.
In other data settings, the baselines will likely be competitive, and thus choosing the right method for learning is dependent on the specific data regime.
We emphasize that our data regime enables scalable learning from easy-to-collect, cheap play data~\cite{mees2022calvin} avoiding the need for expensive task-specific data collection.

\textbf{\textsc{Finding 2:} Continuous latent actions and joint action decoder greatly improve performance in real-world robotics tasks.} 
First, we corroborate the findings by \citet{nikulin2025latent} that continuous latent actions joint training benefit LAM performance.
Unlike \vpt, the other baselines (\lapo, \lapa, and \dynamo) make use of LAMs, as does our method.
We find that we can outperform these methods, likely due to using continuous latent actions in conjunction with jointly training an action decoder.
First, baselines that apply vector quantization \cite{van2017neural} to discretize the latent actions, including \lapo\ and \lapa, perform poorly on continuous control tasks.
In our image-based experiments, \stclam\ achieves an over $3 \times$ improvement in task success rate at $76\%$, compared to \lapo\ and \lapa, which achieve $9\%$ and $20\%$, respectively
(cf.~\Cref{fig:image_results_mw}).
Prior works utilize VQ primarily to simplify the structure of the latent action space which is a reasonable choice for discrete action environments.
We hypothesize that applying quantization to the latent space severely limits the expressivity of the latent actions for fine-grained manipulation tasks.

\begin{table}
\centering
\begin{tabular}{l|cccc}
\toprule
 & \task{Block} & \task{Button} & \task{Microwave} & \task{Slide Pot} \\
\midrule
\bc & 0/10 & 0.5/10 & 1/10 & 0/10 \\
\lapa & 2/10 & 3/10 & 3/10 & 0/10  \\
\vpt & 2.5/10 & 4/10 & 5/10 & 2/10  \\
\stcs & \best{7/10} & \best{8.5/10} & \best{8/10} & \best{4/10} \\
\bce & 7.5/10 & 8/10 & 7.5/10 & 2/10 \\
\bottomrule
\end{tabular}
\caption{\textbf{Real Robot Results}. \stcs\ significantly outperforms baseline methods across all tasks. Each entry sums scores over 10 trials, where partial score or full score is awarded (0.5/1 point). Partial success is, block---reach/pick up; button---near/touch; microwave---touch/close; slide pot---reach/slide.}
\label{tab:real_robot_results}
\end{table}
\begin{figure}[t]
    \centering
    \begin{subfigure}[b]{0.49\linewidth}
        \centering
        \includegraphics[width=\linewidth]{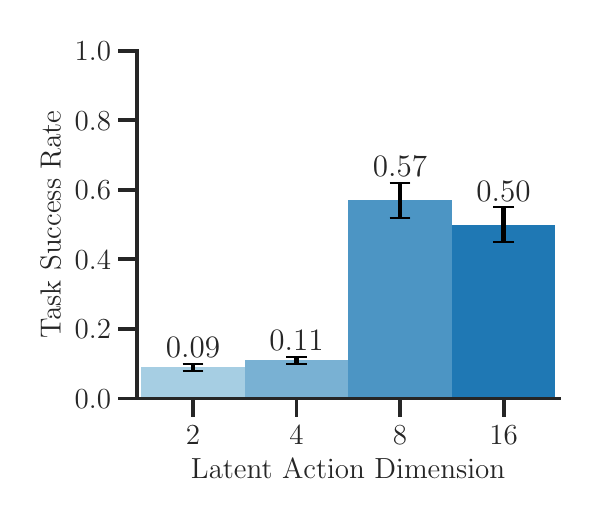}
    \end{subfigure}
    \hfill %
    \begin{subfigure}[b]{0.49\linewidth}
        \centering
        \includegraphics[width=\linewidth]{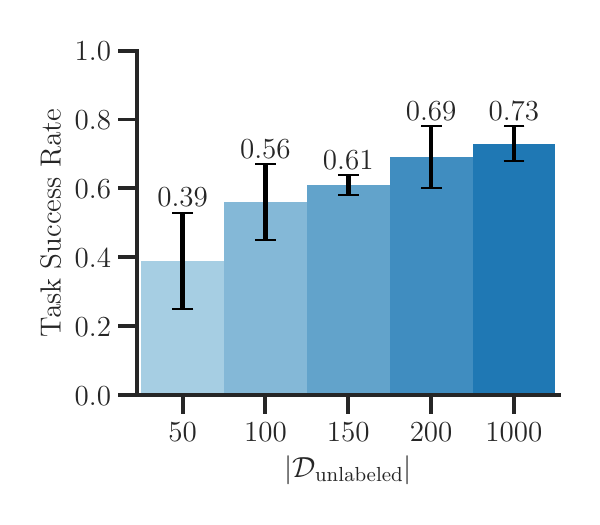}
    \end{subfigure}
    \caption{\textbf{(Left)} Up until a latent dimension of 4, the learned latent action space fails to be useful for IL. 
However, a latent dimension of 8 has sufficient capacity, achieving 57\% success rate on the \texttt{Assembly} task. \textbf{(Right)} CLAM scales with the amount of unlabeled demonstration data. The performance of the downstream policy improves as we annotate more trajectories using the pretrained CLAM.}
    \label{fig:side_by_side}
\end{figure}

A potential issue with using an arbitrary continuous latent action space is grounding actions in the environment.
In discrete settings, LAMs can recover an action space corresponding to a permutation of the ground-truth environment actions~\cite{bruce2024genie}. However, learning this mapping is more challenging for continuous actions.
In Table~\ref{tab:ablate_vq_joint}, we present additional experiments, ablating both the choice of action space (discrete vs.\ continuous) and joint training on MetaWorld tasks. 
Using discrete latent actions, equivalent to \lapo\ \cite{schmidtlearning}, results in an average of $16\%$ task success compared to using continuous latent actions, which achieves $23\%$.
This indicates that even without joint training, continuous actions improve performance.
Furthermore, while joint training does not help much in the discrete latent action case, we see a substantial improvement when coupled with continuous latent actions, achieving $74\%$ average success rate (over $3 \times$ improvement).  

\textbf{Real WidowX Robot Arm Experiments.} 
We evaluate the scalability of CLAM to more realistic applications on a physical WidowX robot arm shown in \Cref{fig:real_robot}~(Left).
Even though $\labeleddata$ comprises of \emph{task-agnostic, mixed expertise play} data, we demonstrate that CLAM is still able to learn to solve the tasks while baseline methods struggle to leverage this data to train a task-specific policy. 
As in the simulated experiments, we find that $\vpt$ is the closest baseline, but still struggles as it can only scale with the limited amount of labeled data, while $\bc$ and $\lapa$ are unable to learn from \emph{play} data entirely. 
Remarkably, CLAM learns a policy to solve new tasks, without having explicitly collected action-labeled, expert demonstrations.
This finding validates prior findings of benefits from action supervision during LAM training, as well as continuous latent action spaces in simulation settings~\cite{nikulin2025latent} and smaller-scale experiments~\cite{yang2025comolearningcontinuouslatent}.

\begin{figure}[t]
    \centering
    \begin{subfigure}[b]{0.49\linewidth}
        \centering
        \includegraphics[width=\linewidth]{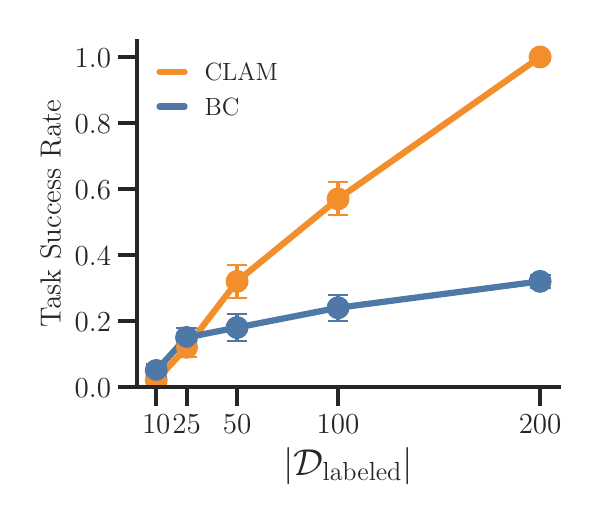}
    \end{subfigure}
    \hfill %
    \begin{subfigure}[b]{0.49\linewidth}
        \centering
        \includegraphics[width=\linewidth]{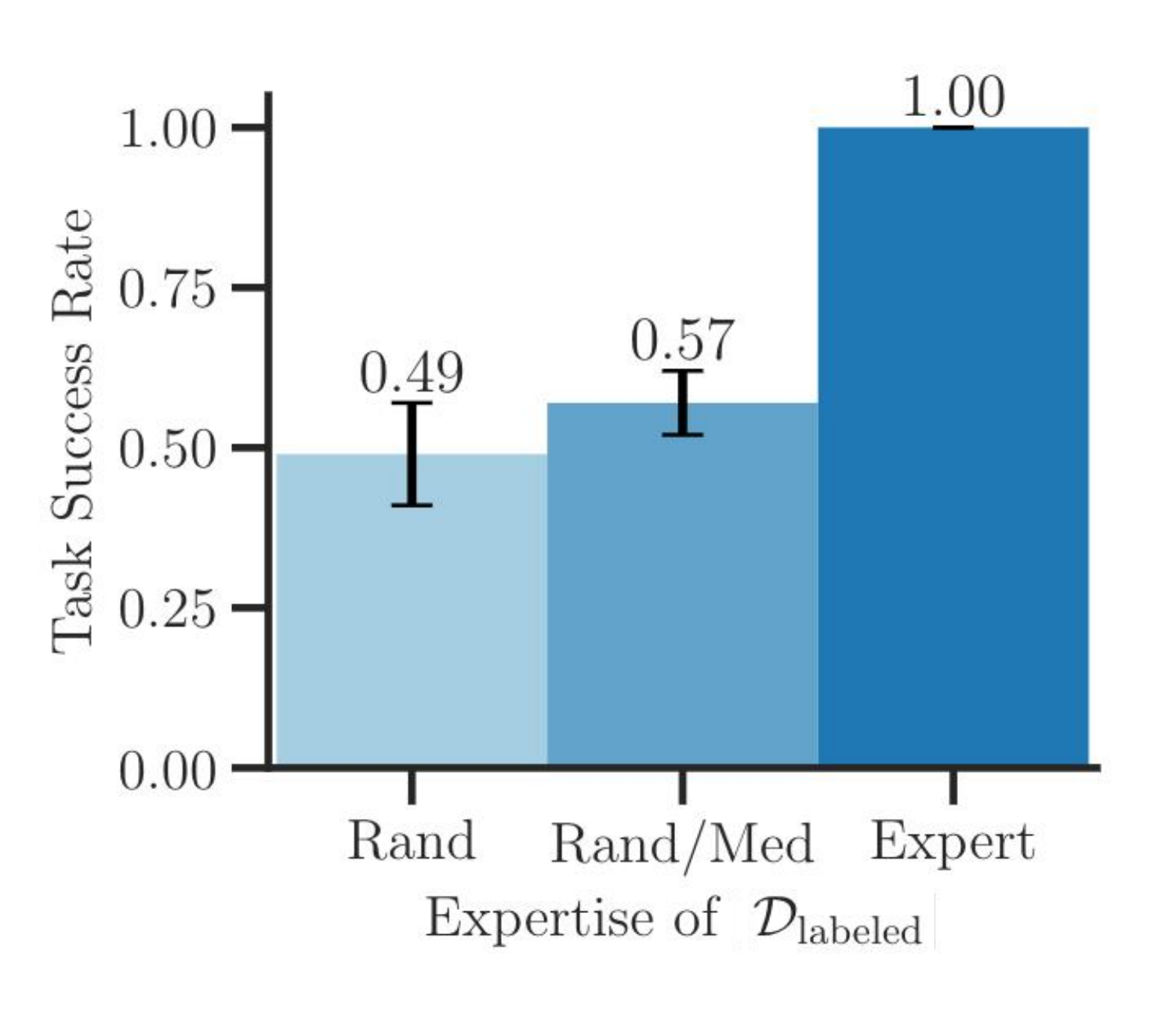}
    \end{subfigure}
    \caption{\textbf{(Left)} We vary the number of labeled trajectories for training the action decoder. 
While BC performance struggles to learn from \emph{non-expert data}, our method improves with more data.
\textbf{(Right)} We also evaluate the robustness of CLAM to data of varying expertise. 
We learn a better policy than BC with the same amount of labeled random trajectories.
With expert data, our method recovers an optimal policy.  
} \label{fig:another_side_by_side}
\end{figure}

\textbf{\textsc{Finding 3:} CLAM successfully learns without ever accessing \emph{action-labeled expert} data in both simulation and real robot experiments.} 
We perform our main-line experiments (\Cref{fig:image_results_mw}) with action-labeled data that is not fully expert.
\lapa\ circumvents the issues of learning a separate action-grounding model by directly fine-tuning the pretrained LAM with an uninitialized action prediction head on $\labeleddata$.
However, if only partially-optimal or play data is available, the fine-tuned BC model struggles to learn a good policy, since it does not label the unlabeled data.
In this regard, \lapa\ is similar to \bc\ with a better initialization as a result of the LAM pretraining, (20\% vs.\ 16\% success rate).
Despite the action-labeled data not being fully optimal, CLAM is still able to achieve high success rates.

To further investigate this capability of our method, we compare data compositions of varying expertise levels.
We filter our offline dataset for 100 trajectories below a predefined threshold return value and manually verify that the policy does not achieve the task and acts randomly. 
We show in \Cref{fig:another_side_by_side}~(Right) that even with \texttt{random} policy data we can achieve similar performance as on the \texttt{random/medium} data regime. 
When \texttt{expert} data is available, we can recover a policy that always solves the task.

%% file: sections/06.2_ablations.tex
\textbf{Latent action dimension directly affects the model's expressivity.} 
In \Cref{fig:side_by_side}~(Left), we vary the latent action dimension $|z| \in \{2,4,8,16\}$ for the MetaWorld \texttt{Assembly} task.
We find that setting $|z|$ to the true action dimension of 4 in this task is insufficient.
Instead, a slight overparameterization is necessary to achieve a reasonable success rate.
This suggests that even when our LAM is capable of learning a viable action space, it might fail at learning the most compact action representation possible.
Further increasing $|z|$ to 16 does not yield any additional gains, likely because once the LAM is capable of learning a viable action space, there is no need for further representational capacity.

\textbf{Latent action policy scales with $|\expertdata|$}. 
In \Cref{fig:side_by_side}~(Right) we demonstrate that CLAM's performance on the \texttt{Assembly} task improves as we increase $|\unlabeleddata|$.
This result suggests that we can improve robot policies without the need for expensive, manually collected expert teleoperated demonstrations. 
We note that the returns start diminishing after a certain number of trajectories, likely because the action decoder's data remains unchanged.

\textbf{Increasing $|\labeleddata|$ improves the action decoder accuracy and downstream performance.} 
In \Cref{fig:another_side_by_side}~(Left) we analyze the effect of varying $|\labeleddata|$ for training the action decoder. 
With only a handful of trajectories, it is difficult to ground the latent actions to the environment explaining the poor performance. 
As we increase $|\labeleddata|$, the learned action decoder becomes more accurate and better generalizes to new unseen states. 
Conversely, BC trained using the same amount of labeled \emph{non-expert} data quickly plateaus in performance and fails to scale to the performance of CLAM, even with up to 100 labeled trajectories.

%% file: sections/07_conclusion.tex
\label{sec:conclusion}
\section{Conclusion}
We proposed CLAM, a scalable solution for learning continuous control tasks from unlabeled robot demonstration data. 
We show with real robot experiments that CLAM, using continuous latent actions and joint training, can learn policies for unseen tasks without requiring any expert teleoperated demonstrations.
We hope CLAM further advances the scalable training of robot policies from action-less data, alleviating the need for expensive, manual data collection.

%% file: sections/08_limitations.tex
\label{sec:limitations}

\textbf{Generalizing to real-world videos.}
To train foundation policies from in-the-wild Internet scale video data, a number of additional challenges need to be addressed.
In-the-wild internet videos might contain visual artifacts such as motion blur, low resolution, and compression artifacts, which might have to be addressed with existing computer vision techniques.
Another challenge is the embodiment gap between human videos and the target robot embodiment.

\textbf{Embodiment gap.} 
The human-robot embodiment gap is directly addressable with video conversion methods such as Phantom~\cite{lepert2025phantom} and Masquerade~\cite{lepert2025masquerade}.
Alternatively, prior works such as CrossFormer \cite{Doshi24-crossformer} fine-tune action heads for different embodiments, but are limited to a fixed set of embodiments seen during training.
For future work, we aim to apply the action-less learning capabilities of CLAM to the cross-embodiment problem, to allow learning for various robot embodiments and the human embodiment in a scalable way.

%% file: sections/appendix/appendix.tex
\begin{table}[H]
\vspace{-12pt}
\centering
\begin{tabular}{lc}
\toprule
\textbf{Hyperparameter} & \textbf{Value} \\ 
\midrule
Num updates                & 500,000       \\
Train action decoder every & 2             \\
Action Decoder batch size  & 128           \\
Action Decoder loss weight & 1             \\
Action Decoder hidden dim  & [1024, 1024, 1024] \\
Action Decoder embedding dim & 512         \\
Reconstruction loss weight & 1             \\
Latent action dim          & 16            \\
Context len                & 2             \\
Embedding dim              & 128           \\
\bottomrule
\end{tabular}
\vspace{-0.5em}
\caption{CLAM Pretraining Hyperparameters}
\label{tab:clam_hparams}
\end{table}

\begin{table}[H]
\vspace{-15pt}
\centering
\begin{tabular}{lc}
\toprule
\textbf{Hyperparameter} & \textbf{Value} \\ 
\midrule
Num encoder layers        & 3        \\
Num decoder layers        & 3        \\
Model dimension           & 256      \\
Feedforward dimension     & 2048     \\
Num attention head        & 4        \\
Dropout                   & 0.1      \\
Pre norm                  & False    \\
Feedforward activation    & \texttt{GeLU}     \\
Position Encoding         & Learned  \\
\bottomrule
\end{tabular}
\vspace{-0.5em}
\caption{Transformer CLAM Model Hyperparameters}
\label{tab:transformer_clam_hparam}
\end{table}

\begin{table}[H]
\vspace{-15pt}
    \centering
    \begin{tabular}{@{}l@{}ccc@{}}
        \toprule
        & \textbf{MuJoCo} & \textbf{MetaWorld} & \textbf{Real Robot} \\
        \midrule
        $\mathcal{D}_{\text{unlabeled}}$ & 1000 & 1000 & $\sim$500 \\
        $\mathcal{D}_{\text{labeled}}$ & 50 & 50 & 50 \\
        $\mathcal{D}_{\text{unlabeled\_expert}}$ & 20 & 20 & 30/50 \\
        \bottomrule
    \end{tabular}
    \caption{Data Splits (\# trajectories)}
    \label{tab:data_splits}
    \vspace{-15pt}
\end{table}

\begin{table}[H]
\centering
\begin{tabular}{l|cc}
\toprule
Domain & Action Dim \\
\midrule
Hopper                         & 3 \\
HalfCheetah                    & 6 \\
MetaWorld (all tasks)          & 4 \\
Real WidowX (all tasks)        & 7 \\
\bottomrule
\end{tabular}
\vspace{-0.5pt}
\caption{Ground-Truth Action Dimensions 
}
\label{tab:action_spaces}
    \vspace{-10pt}
\end{table}